\documentclass[12pt, onecolumn, letterpaper]{article}
\makeatletter

\usepackage{amsmath}
\usepackage{amssymb}
\usepackage{algorithm,algpseudocode}
\usepackage[mathscr]{euscript}
\usepackage{hyperref}
\usepackage{graphicx}
\usepackage{caption}
\usepackage{subcaption}
\usepackage{color,soul}

\date{September 18, 2021}
\title{\LARGE \bf Design and Development of a Remotely-enabled Modular Release Mechanism for Autonomous Underwater Vehicles\thanks{As of May 30, 2023, the work described herein is covered under US Patent 11,661,153}}

\author{Demetrious T. Kutzke\thanks{Corresponding author: Demetrious T. Kutzke~({\tt\small Demetrious.Kutzke1@navy.mil})}
$^{\ddagger}$, Gustavo E. Miranda L\'{o}pez$^{\ddagger}$, \\Robert J. Herman$^{\ddagger}$, and Harryel Philippeaux%
\thanks{The authors are with the Naval Surface Warfare Center Panama City Division, Panama City, FL 32407, USA.}%
}

\begin{document}
\maketitle

\begin{abstract}
We introduce a launch device, called the remotely-enabled modular release mechanism, to augment rapid testing and prototyping of cooperative autonomy maritime applications by facilitating autonomous deployment of an autonomous underwater vehicle (AUV) from an autonomous surface vessel (ASV). While we focus our development on a specific application of deploying an AUV from a catamaran style ASV, the release mechanism can be adapted to different deployable objects and towing vehicles, such as buoys and sensors for oceanographic surveys or mono-hull ASVs. In this paper we explore a number of hardware and software design considerations to facilitate ease of integration with existing maritime autonomy systems. We expound on bench tests and in-water tests used to explore the utility of the release system and diagnose system issues. Additionally, we make a first-principles argument, based on a hydrodynamics physics model, for assured deployment that is virtually independent of sea state, making the release system a suitable alternative for different maritime applications in varying environmental conditions.
\end{abstract}
\section{Introduction}
Cooperative autonomy solutions show promise
for accomplishing traditionally challenging tasks such as cleaning and containing oil spills,
sensor network deployments, medical supply deliveries, package deliveries,
and natural disaster recovery systems \cite{song2020care,song2013adaptive,rizk2019cooperative,ackerman2018medical,limosani2018robotic,jorge2019survey,Gatteschi2015_DroneDeliverySystems,yukun2018_UnderwaterVehicleForMarineLife,gonzalez2020_AUVCollaboration,utne2019}.
Cooperative autonomy here refers to multiple robots or autonomous agents
working in a cooperative fashion to accomplish a task or a number of subtasks.
These applications often require the coordination of distinct tasks across heterogeneous agents or robot teams. 
This means that often one agent's schedule is coupled to another agent's schedule through the task assignment.
These cross schedule dependencies present significant
challenges to develop task allocation and scheduling algorithms and represent an even more
challenging task to satisfy in practice \cite{korsah2012xbots,kutzke2021generosity,zitouni2020towards,chen2019cost}.

In maritime applications, for example, transporting, docking, and deploying
an autonomous underwater vehicle (AUV) from a manned or
autonomous surface vehicle (ASV) are common tasks that are required to perform
some of the aforementioned cooperative autonomy applications. Developing schedules
around these tasks falls broadly under the
multi-robot task allocation set of problems and doing so in practice can be very challenging, particularly for underwater robotics since
communication assumptions breakdown and require alternative collaboration techniques \cite{gerkey_taxonomy,Kalwa2009,birk2011,real2016,Kalwa2015}.

Our contribution is the development of a 
technology that augments rapid prototyping of cooperative autonomy solutions
that need to autonomously deploy something, be it an AUV, buoy, or a sensor from an ASV. It is appropriate to think of the device, which
we call the remotely-enabled modular release mechanism or (RM)\textasciicircum2
as an experimental apparatus, the use of which allows development
teams to quickly test and integrate advanced autonomy without
any additional development time to acquire or build a custom solution.

Two critical ideas form the basis of this work: First,
experimental autonomy requires numerous technology enablers, not the least
of which is algorithm and vehicle specific hardware to ``prove"
or provide an experimental basis for certain theoretical algorithms. Second,
without the experimental apparatus in place, there is no clear path
that goes directly from theory and numerical simulation to commercial application.

Cost and design considerations must be considered as well, since cooperative autonomy solutions typically begin with
a bulk of the research focused on the theoretical and numerical developments of the underlying algorithms and
do not permit extensive experimental budgets. Moreover, there is little incentive for academia or government
to invest in the validation of ``proven'' cooperative solutions, since there is little chance of publication success and
little applicability to the needs of government. Industry cannot justify investing in theoretical algorithms, if
there is no clear path to commercial success. To mitigate these challenges which often leave technically
rigorous and innovative theoretical results relegated to simulation results, early stage science and technology
projects need affordable, low-cost alternatives for proof-of-concept and experimental testing of theoretical autonomy projects. 

The release mechanism we introduce in this paper
offers just that---a low cost alternative that allows rapid testing and prototyping of cooperative autonomy
for maritime applications. While we focus our development on a specific application of deploying a AUV from an ASV, the release mechanism can be adapted to different ``deployable'' objects and towing vehicles.
We are primarily concerned with a number of hardware design considerations and software development 
to facilitate compliance with IoT industry standards. In summary, our main contributions to the field of launch and recovery of unmanned systems are

\begin{enumerate}
\item An affordable, low-cost alternative for deploying AUVs from catamaran style ASVs
\item An argument for assured release under varying sea state conditions
\item A control module written in the robot operating system for easy integration with existing autonomy software
\end{enumerate}

%\nomenclature[A]{IoT}{Internet of things}

The paper is outlined as follows: In Section~\ref{sec:related_work} we discuss the historical development of release
mechanisms for unmanned vehicles. In Section~\ref{sec:mechanical_design_considerations} we discuss
the hardware and environmental factors affecting design implementation. In Section~\ref{sec:iot_architecture_considerations}
we discuss industry standards for Internet of Things data interfaces that affect software design considerations.
Section~\ref{sec:experimental_testing} introduces the experimental testing that was conducted to determine
the efficacy of the release mechanism. We conclude in Section~\ref{sec:conclusion} with a discussion on 
the results and future work.
\section{Related Work}
\label{sec:related_work}
Many efforts have been dedicated to launch and recovery devices of AUVs from surface vessels and underwater vessels such as submarines \cite{wigley2018,fiaz2018intelligent,ead2007apparatus,ansay2008pre,banerjee1995deployment,patrick2016mechanisms,pugi2018redundant}. Often launch and recovery devices use a cable or winch mechanism to both deploy and recover the AUV onto the vessel. For example, Renilson describes a proposed solution to recover a AUV from a submarine that does not require the use of torpedo tubes or slow maneuvering by the submarine to allow the AUV to overtake the submarine. Instead, Renilson proposes an innovative approach that utilizes a cable system to recover the AUV via a cable that drops from the AUV and is captured by the submarine \cite{renilson2014simplified}. Alternatively, there exist numerous studies on the hydrodynamics of launching AUVs from tube-like structures. Zhang et al. provide a detailed account of the complexities of deploying a AUV from a torpedo tube, including requirements on launch thrust requirements throughout the deployment process \cite{zhang2021_Hydro}. Much of the focus on docking systems assumes design considerations such as a fixed dock or certain sea state considerations for tethered systems \cite{hobson2007development,allen2006autonomous}. To accommodate for complex docking situations, some have propounded sophisticated algorithmic approaches. Trsli{\'c} et al. have considered an algorithm-focused approach that utilizes fuzzy logic for localization control for docking remotely operated vehicles (ROVs) to floating docks on offshore drilling platforms \cite{trslic2020neuro}. They suggest that due to limited resource constraints on AUVs, alternative methods must be utilized to predict heave motion of docking stations suspended on the surface. Landstad et al. utilize dynamic positioning control to predict the wave motion of the surface vessel that the remotely operated vehicle can use to match and predict when to dock with the surface vessel \cite{Landstad2021_DynamicPositioning}. More recently Zhao et al. have provided a detailed study on a hydrodynamic model of the coupling between an ROV and an ASV system. This coupling is of supreme importance since as the ROV approaches the ASV for docking or capture, it becomes increasingly susceptible to the wave drive motion of the AUV, making docking an extremely challenging task from a control perspective \cite{zhao2021_LRD}. Meng et al. demonstrate the ``capture rod'' method, which utilizes a rod or cable that extends from a docking station. The AUV then utilizes an acoustic signal for homing \cite{meng2019_UnderwaterDocking}. These methods vary in sophistication, since recovery devices are far more sophisticated and require a great deal more effort to recover a AUV than to release or deploy a AUV system. Some have approached the problem from an operational research perspective or even a machine learning perspective to handle localization in the vicinity of the capture rod \cite{cris2021_TaskPriority,anderlini2019_DockingControlAUVReinforcement}. Allotta et al. have also explored docking considerations for underwater wireless recharging stations. They too encounter similar hydrodynamic issues when looking at docking robotics platforms on charging stations subject to undersea currents. 

The preceding examples provide substantial inspiration for developing a release system that assumes similar or simplified assumptions. We acknowledge that this problem is trivial to a certain extent; however, previous attempts at combining both launch and recovery systems have proven very complex, costly, and challenging to adapt to legacy systems. In our work, we are not concerned with recovery, since it is assumed that whatever is being deployed can either be recovered by hand or will operate in the ``persistent'' autonomy space, which is long duration missions that require little to no human intervention. We also consider that the release mechanism should not be invasive to the towing system, which according to Kouriampalis et al. can be quite challenging from an operational and ship design perspective to accommodate for launching and recovering AUV systems \cite{kouriampalis2021_OperationalEffects}. Since we are targeting those end users who would benefit from a so-called ``plug and play'' device, then we do not want to burden the developer with significant changes to the available towing vessel.

Sarda and Dhanak introduce a design concept very similar to our release system. Indeed, the similarities are striking \cite{sarda2014}. There are some important distinctions, however. First, they utilize a release method that requires the nose and tail of the AUV to enter the water at the same time. During the deployment phase the AUV is lowered from the payload tray of the towing vessel, utilizing two winches attached below the payload tray of a wave adaptive modular vessel (WAM-V). The AUV is suspended from underneath the WAM-V during transit, which introduces additional control problems for the autonomy software. We take an alternative approach and utilize the REMUS vehicle's neutral buoyancy to tow the AUV on the surface between the catamarans. We utilize a very simple PVC style housing to contain the AUV during transit. While the winch system of Sarda and Dhanak is a unique design concept, we argue that winch systems add additional weight, resource requirements (such as additional power sources), and reduce modularity by placing restrictions on the towline. Second, Sarda and Dhanak have invested significant effort into modeling and simulation of the multi-step process of launch and recovery. Modeling and simulation (M\&S) is important when considering a process as complex as recovery of the AUV onto the towing vessel. Their efforts are admirable and quite impressive but beyond the scope of our discussion, since we have had the advantage of a decidedly less complex scenario of launch or deployment, exclusively. Third, to our knowledge, Sarda and Dhanak proposed but never implemented the launch and recovery system they propounded in their work. To that end, they have not tested the system in an experimental environment, whereas we have designed, built, and tested our release mechanism in both laboratory and real-time test events. 

\section{The (RM)\textasciicircum2 System Description}
\begin{figure*}[t]
\centering
\begin{subfigure}[t]{0.49\textwidth}
\centering
\includegraphics[width=1.0\textwidth]{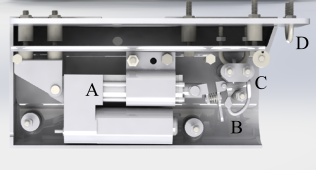}
\caption{Side view of internal actuator and spring with hook system}
\label{fig:side_cut}
\end{subfigure}
\begin{subfigure}[t]{0.49\textwidth}
\centering
\includegraphics[width=1.0\textwidth]{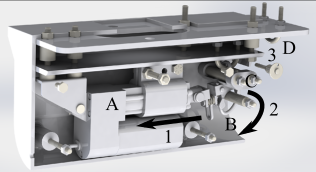}
\caption{Angled view of internal actuator and spring with hook system}
\label{fig:side_cut_angle}
\end{subfigure}
\caption{(Color online) Renderings of the release mechanism's internal components}
\label{fig:internal}
\end{figure*}

The system works according to the architecture described in Fig.~\ref{fig:architecture} and the physical system sketch shown in Fig.~\ref{fig:main_design}. The user specifies a `DEPLOY\_POSITION' in the form of a comma delimited latitude and longitude pair, or a `DEPLOY\_TRIGGER,' which is a string alias, in the catkin script for either the $\textproc{RM2Node}$ directly or in the global catkin configuration for the user's software. The $\textproc{RM2Node}$ subscribes to `POSITION' which is updated by the navigation software of the vehicle, typically published in a so-called local coordinate system that maps a latitude and longitude to an origin and all further position updates are in the local coordinates. Internally, $\textproc{RM2Node}$ converts all coordinates to local coordinates. $\textproc{RM2Node}$ handles conversion between coordinate systems utilizing the library MOOSGeodesy \cite{moosgeodesy}. Once the $\textproc{RM2Node}$ registers an update on its subscribers, it checks if the difference between the current position $\vec{x}'$ and the desired deployment location $\vec{x}$ is within the tolerance distance $\delta$ or if the trigger has been updated to $\textit{true}$. If either of these conditions is true, then the node triggers a serial connection with the low-level Arduino controller, sending a single ASCII encoded character `D' to signal that the controller should actuate the release mechanism. The physical release mechanism consists of a linear actuator that drives a pin within the housing. A three dimensional rendering is shown in Fig.~\ref{fig:internal}. In Fig.~\ref{fig:side_cut}, ``A'' marks the linear actuator, ``B'' is the hook or shackle, ``C'' is the pin that maintains the hook in the locked position, and ``D'' is the loop that ensures a consistent angle between the tether and the hook at all times. Without the pin, the hook will fall open due to gravity. The tether runs between ``C'' and ``D.'' The pin is attached to a spring that is pulled using the actuator. In Fig.~\ref{fig:side_cut_angle}, there are three main events that take place in succession of control. The first event, labeled as ``1'' in Fig.~\ref{fig:side_cut_angle}, pulls the pin in the direction of the arrow. The second event occurs once the pin is pulled, allowing the hook to move in the direction of arrow ``2.'' In the third event, labeled as ``3,'' the tether slides through the U-bolt and passes back through the handle of the AUV which is held between the pontoons within the PVC housing. This is to ensure that the tether remains with the WAM-V and does not foul the propellers on the AUV once released. 

\begin{figure}[t]
\centering
\includegraphics[width=0.45\columnwidth]{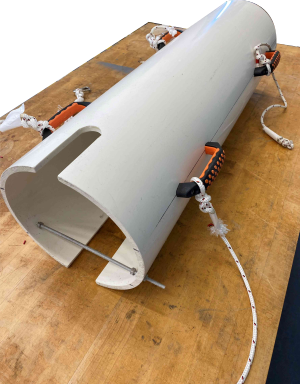}
\caption{(Color online) PVC housing for the REMUS 100. The picture is oriented to show the nose portion of the housing}
\label{fig:housing}
\end{figure}

An image of the PVC housing is shown in Fig.~\ref{fig:housing}. The PVC is a commercial off the shelf product, $9.5~\text{in}$ or $0.241~\text{m}$ in diameter. It is $0.838~\text{m}$ in length. A $3.81~\text{cm}$ notch was cut in the nose portion of the PVC to accommodate the tether that loops through the handle of the REMUS 100. The notch down the underneath portion of the PVC is to permit the ballast of the REMUS to slide in as well. The REMUS 100 will power on to an active state, which starts mission execution, once a magnet activates the side panel of the vehicle. A magnet is attached to the interior of the PVC toward the aft part of the housing, which physically touches the side of the AUV as the AUV exits from the housing. A summary of the (RM)\textasciicircum2 system specifications is shown in Table~\ref{table:specs}.

\begin{table}[!t]
\renewcommand{\arraystretch}{1.1}
\caption{(RM)\textasciicircum2 physical dimensions and specifications}
\label{table:specs}
\centering
\begin{tabular}{cc}
\hline
Specification&Numerical value\\
\hline\hline
Physical dimensions&$(35.56~\text{l}\times16.51~\text{h}\times12.7~\text{w})~\text{cm}$\\
Total mass&3.95~kg\\
Linear actuator force&1468~N\\
Linear actuator stroke&5.08~cm\\
\hline
\end{tabular}
\end{table}

\section{Mechanical Design Considerations}
\label{sec:mechanical_design_considerations}
\begin{figure*}[t]
\centering
\begin{subfigure}[t]{0.32\textwidth}
\centering
\includegraphics[width=1.0\textwidth]{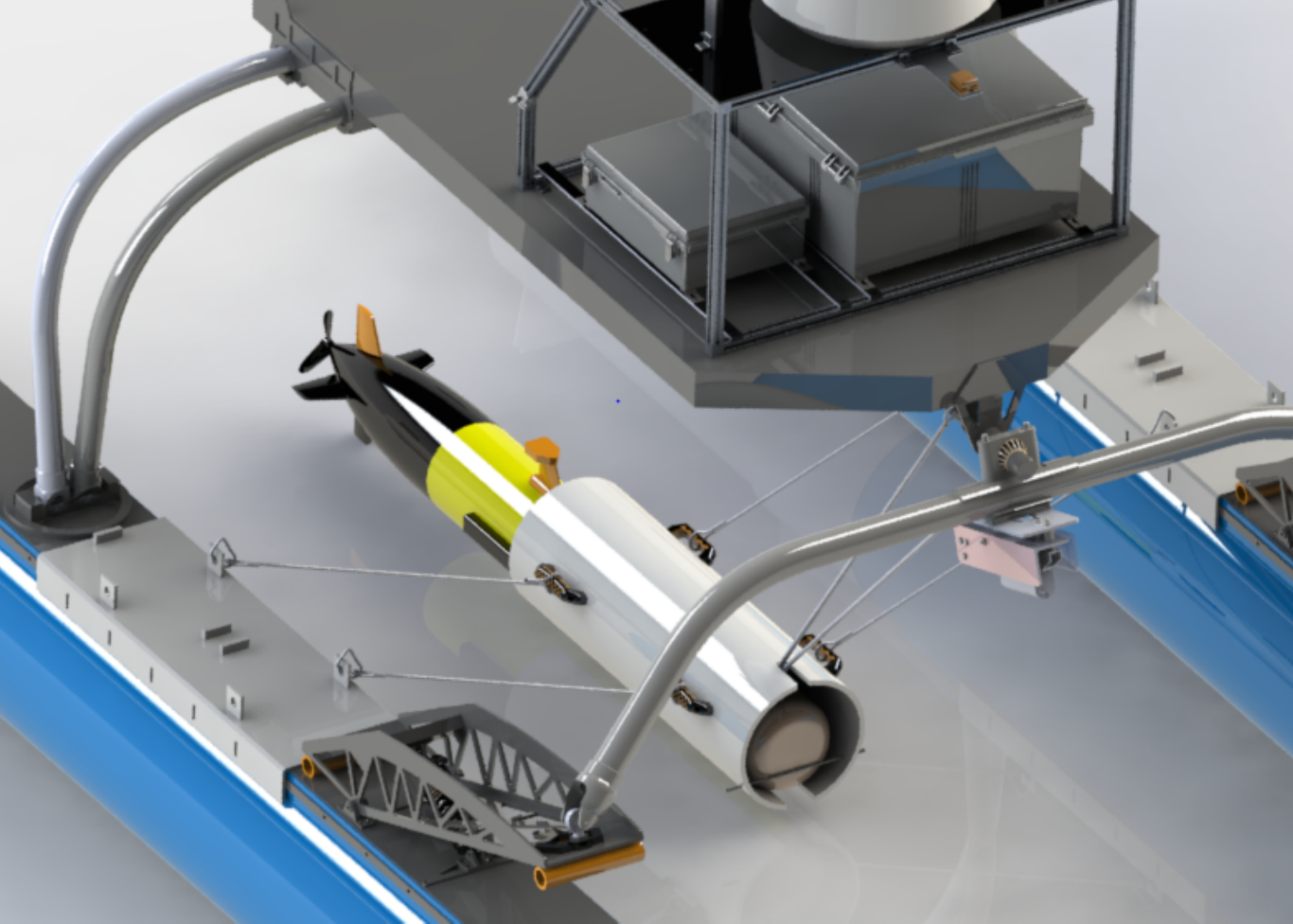}
\caption{Top view}
\label{fig:main_design_top_view}
\end{subfigure}
\begin{subfigure}[t]{0.32\textwidth}
\centering
\includegraphics[width=1.0\textwidth]{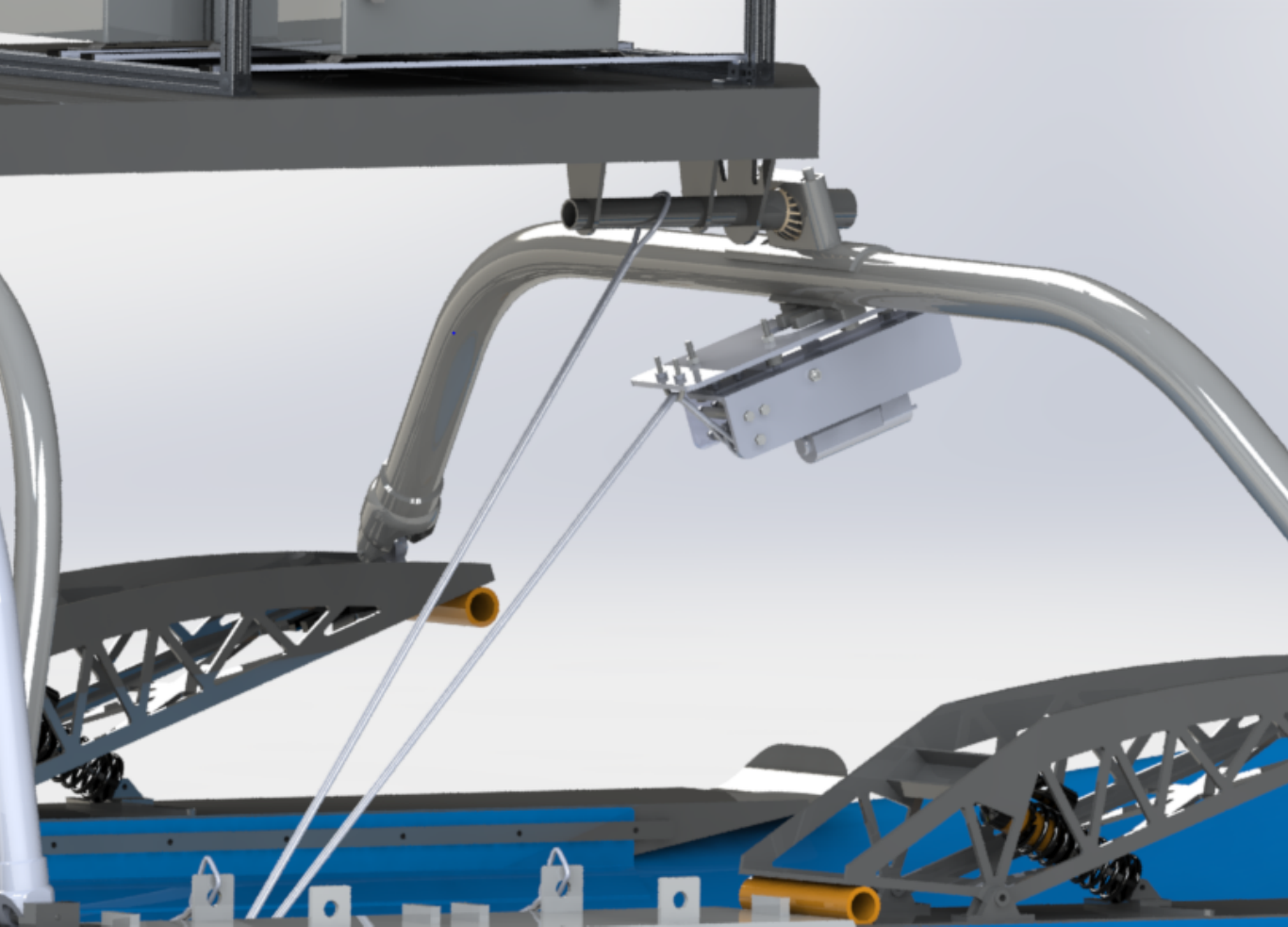}
\caption{Side view}
\label{fig:main_design_side_view_zoomed_out}
\end{subfigure}
\begin{subfigure}[t]{0.32\textwidth}
\centering
\includegraphics[width=1.0\textwidth]{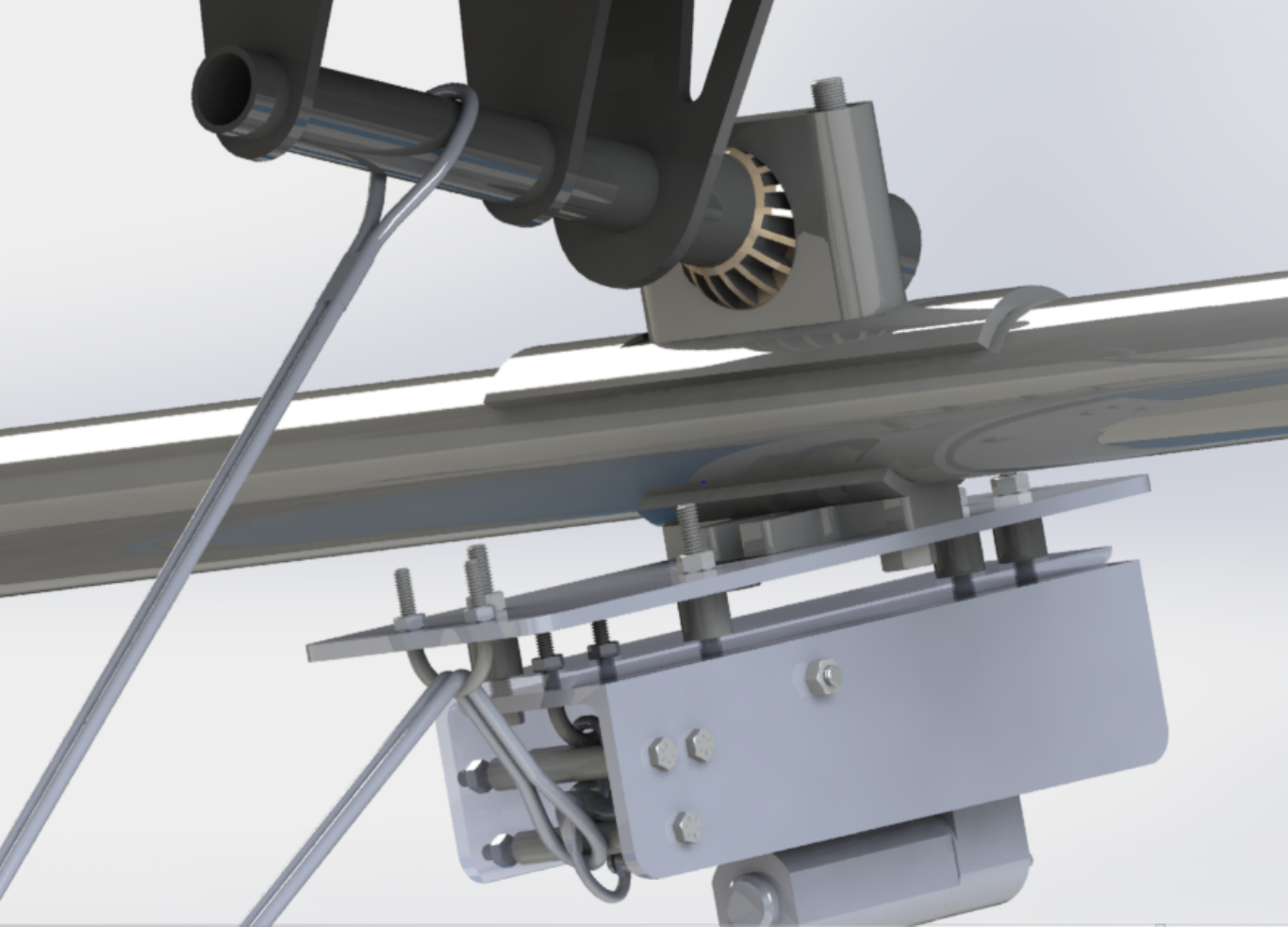}
\caption{Side view close up}
\label{fig:main_design_side_view}
\end{subfigure}
\caption{(Color online) Artistic renderings of (RM)\textasciicircum2 configurations}
\label{fig:main_design}
\end{figure*}

The primary mechanical design aspects considered when first developing the system were size, weight, towing strength, and design for maneuverability. An artistic rendering of (RM)\textasciicircum2 is shown in Fig.~\ref{fig:main_design}. The device was constructed to interface with the WAM-V\thanks{WAM-V is a registered trademark of Marine Advanced Robotics, Inc.} 16 Autonomous Surface Vessel, manufactured by Marine Advanced Robotics, Inc \cite{wamv}. The vehicles are manufactured with flexible pontoons that adapt to varying sea state conditions, making the WAM-V a suitable vessel for maritime autonomy research, since they are light-weight, easily transportable, easy to assemble, and modular for adjusting payloads. The specific vessel chosen is inconsequential to the development of (RM)\textasciicircum2, however, since it is a relatively simple task to adapt (RM)\textasciicircum2 by changing the face plate (see Fig.~\ref{fig:main_design_side_view}) to accommodate alternative vessels. The plate is bolted to the existing hinge that fixes the payload tray to the catamaran arms. 

\begin{figure}[ht]
\centering
\includegraphics[width=0.46\columnwidth]{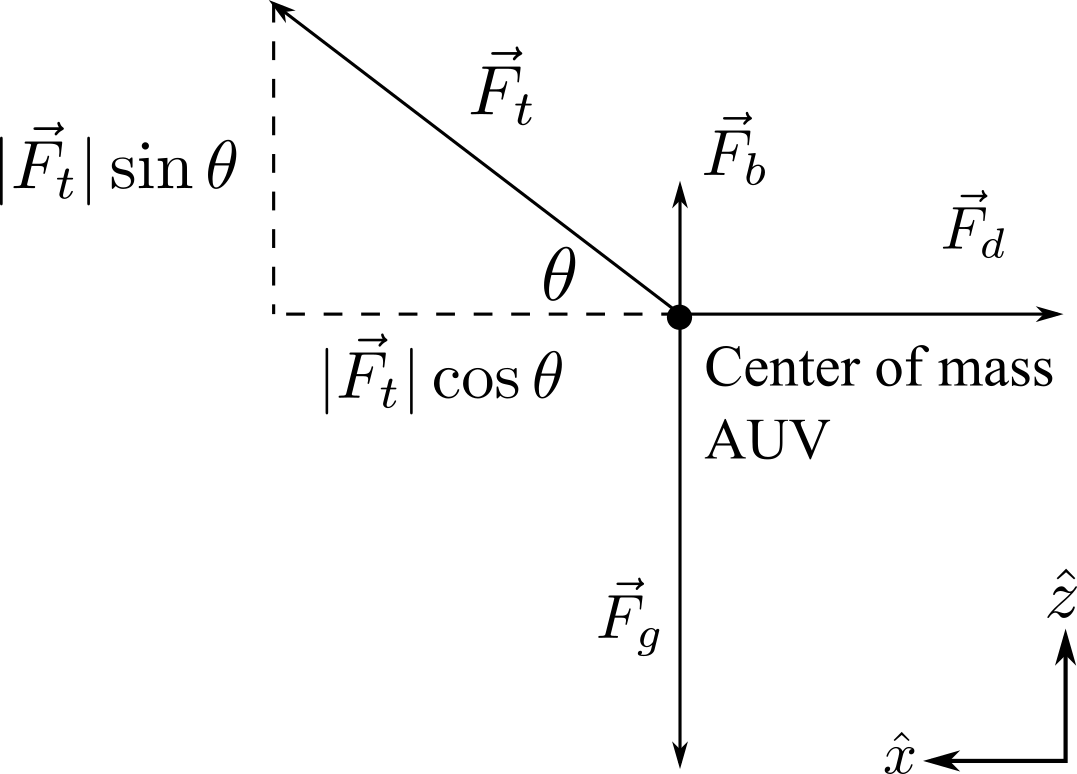}
\caption{Free body diagram of the equilibrium forces incident on the AUV under tow}
\label{fig:free_body}
\end{figure}

The size of (RM)\textasciicircum2 was constrained to minimize the occupied space on the vessel, since we did not want to have to modify the vehicle directly or utilize precious payload space for external sensing and attachments, all of which would degrade modularity of the vessel. Moreover, care was taken to avoid any fouling that might occur due to the tethers attached to the pontoons and the tow cable attached directly to the vehicle. Fig.~\ref{fig:main_design_top_view} shows how a simple tether system is used to maintain the position of the AUV within the PVC housing between the pontoons of the WAM-V. The tethers that attach the housing to the pontoons are nylon braided rope and constrained to be about $2/3~\text{m}$ in length with a diameter of approximately $1.27~\text{cm}$. (RM)\textasciicircum2 is positioned fore on the towing vehicle rather than aft to mitigate fouling in the motors. The weight of (RM)\textasciicircum2 was constrained to occupy a negligible amount of payload weight, which could affect maneuverability and potentially compromise the integrity of the vessel. (RM)\textasciicircum2 measures approximately $3.9~\text{kg}$ in assembled mass, compared to the overall payload capacity of the WAM-V which is $113~\text{kg}$ using the four battery configuration. The towing strength was explored to ensure that (RM)\textasciicircum2 could at least release a towed load equivalent to that of the tow strength of the tether. Practically speaking, the tow strength is bounded above by the tow strength of the vessel, which through experiment we determined was far greater than the towed unmanned underwater vehicle or approximately $110~\text{N}$ of tension in the towline. This tension was approximated using a simplistic free body diagram shown in Fig.~\ref{fig:free_body}. 

There are four primary forces acting on the AUV: The drag force, labeled as $\lvert\vec{F}_{d}\rvert$, acts on the AUV in the $-\hat{x}$ direction and is proportional to the effective cross sectional area acted on by the fluid; the gravitational force, labeled as $\lvert\vec{F}_{g}\rvert$; the force due to tension in the towline, labeled as $\lvert\vec{F}_{t}\rvert$; and finally the buoyant force that acts opposite to the direction of the gravitational force. The buoyant force is proportional to the volume of the fluid displaced by the submerged portion of the AUV's hull. Assuming there is no net force on the AUV and the WAM-V moves with speed $\upsilon$ in the $\hat{x}$ direction, then it is a relatively straightforward task to calculate the tension in the towline. For the forces in the $\hat{x}$ direction,

\begin{equation}
\lvert\vec{F}_{t}\rvert\cos{\theta} - \frac{\rho}{2}C_{D}\sigma\upsilon^2 = 0,
\end{equation}

\noindent and 

\begin{equation}
\lvert\vec{F}_{t}\rvert = \frac{\rho}{2\cos{\theta}}C_{D}\sigma\upsilon^2.
\label{eqn:tension}
\end{equation}

In (\ref{eqn:tension}), $\rho$ is the density of the fluid medium, in this case, surface seawater which is approximately $1020~\text{kg/m}^3$; $\theta$ is approximately $45$\textdegree; $C_{D}$ is the drag coefficient, which utilizing a half-sphere to approximate the nose of the AUV is $0.42$; $\sigma$ is the effective surface area, and in this case $\sigma\approx \pi r^2/2 = 0.057~\text{m}^2$; and finally $\upsilon\approx 2.5~\text{m/s}$. Substituting these values in (\ref{eqn:tension}) yields approximately $109~\text{N}$ of force in the towline. The towline used was rated for nearly $2\times10^3~\text{N}$ safe load, indicating that the total tension in the line under tow is significantly below the rated tow capacity. Since there is no net force, and the AUV is not completely suspended but only kept on the surface between the pontoons then the $\hat{z}$ forces can be neglected without concern for any payload effects.

\subsection{Total Cost}
Since the total funding level for an early stage development effort is usually
in the range of $10^{4}$--$10^{5}\text{K}$ USD, we considered commercial off-the-shelf
components for (RM)\textasciicircum2 that did not exceed $1\text{K}~\text{USD}$ in assembled cost to ensure that the acquisition of (RM)\textasciicircum2 was nearly negligible when compared to the overall funding level \cite{link2010government}.

\subsection{Environmental Conditions}
\begin{figure*}[t]
\centering
\begin{subfigure}[t]{0.33\textwidth}
\centering
\includegraphics[width=1.0\textwidth]{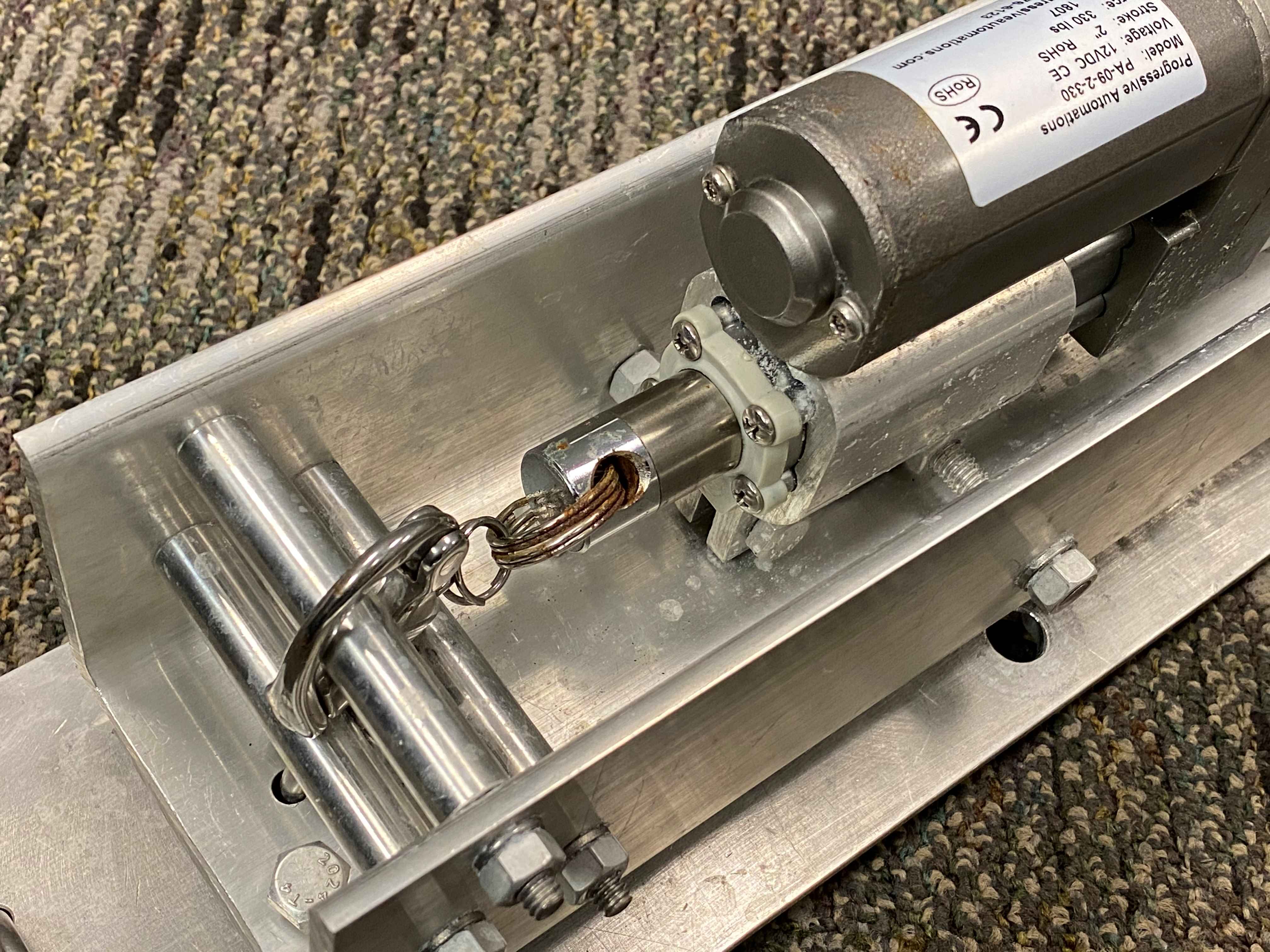}
\caption{}
\label{fig:corrosion1}
\end{subfigure}
\begin{subfigure}[t]{0.33\textwidth}
\centering
\includegraphics[width=1.0\textwidth]{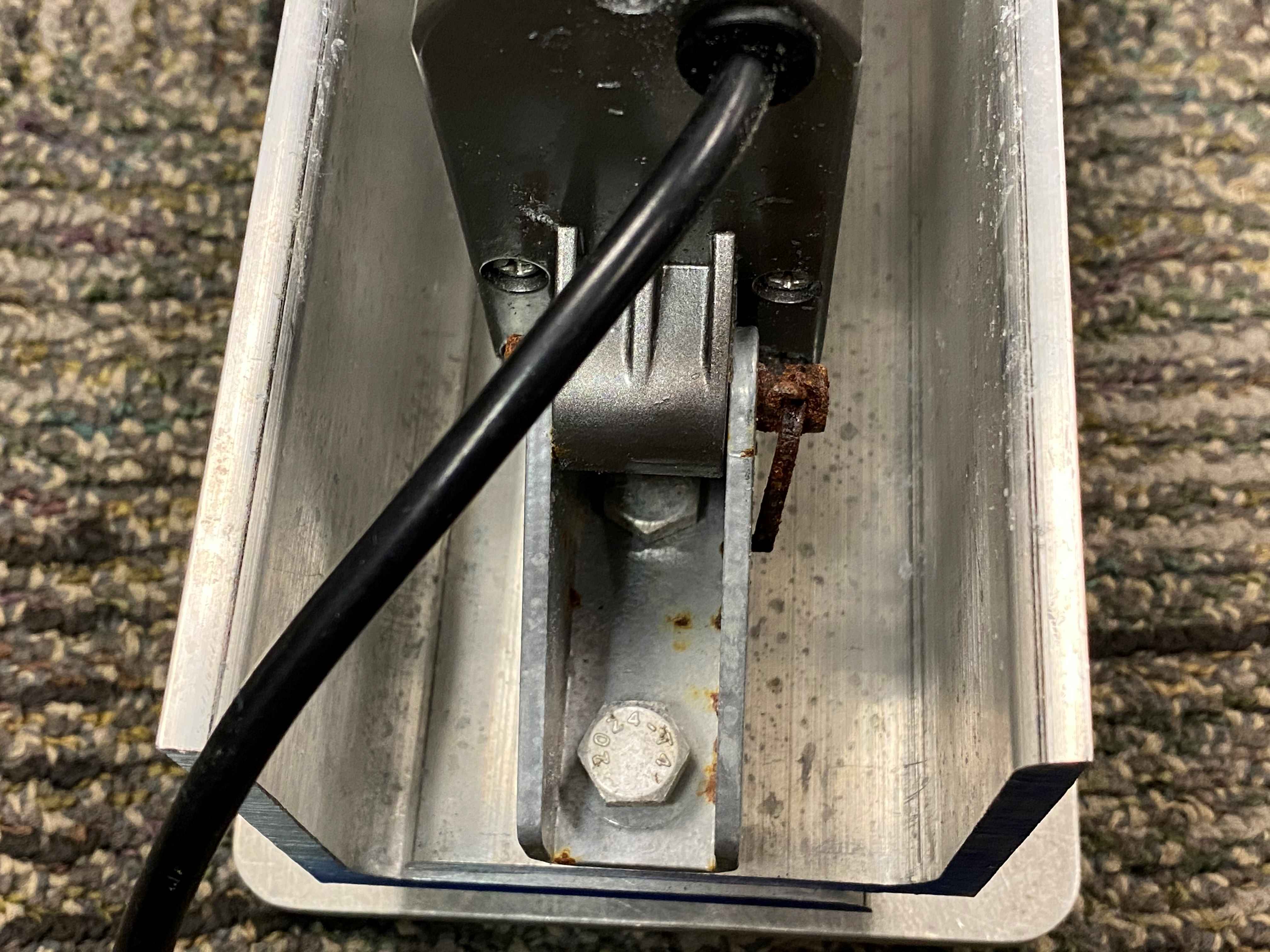}
\caption{}
\label{fig:corrosion2}
\end{subfigure}
\begin{subfigure}[t]{0.33\textwidth}
\centering
\includegraphics[width=1.0\textwidth]{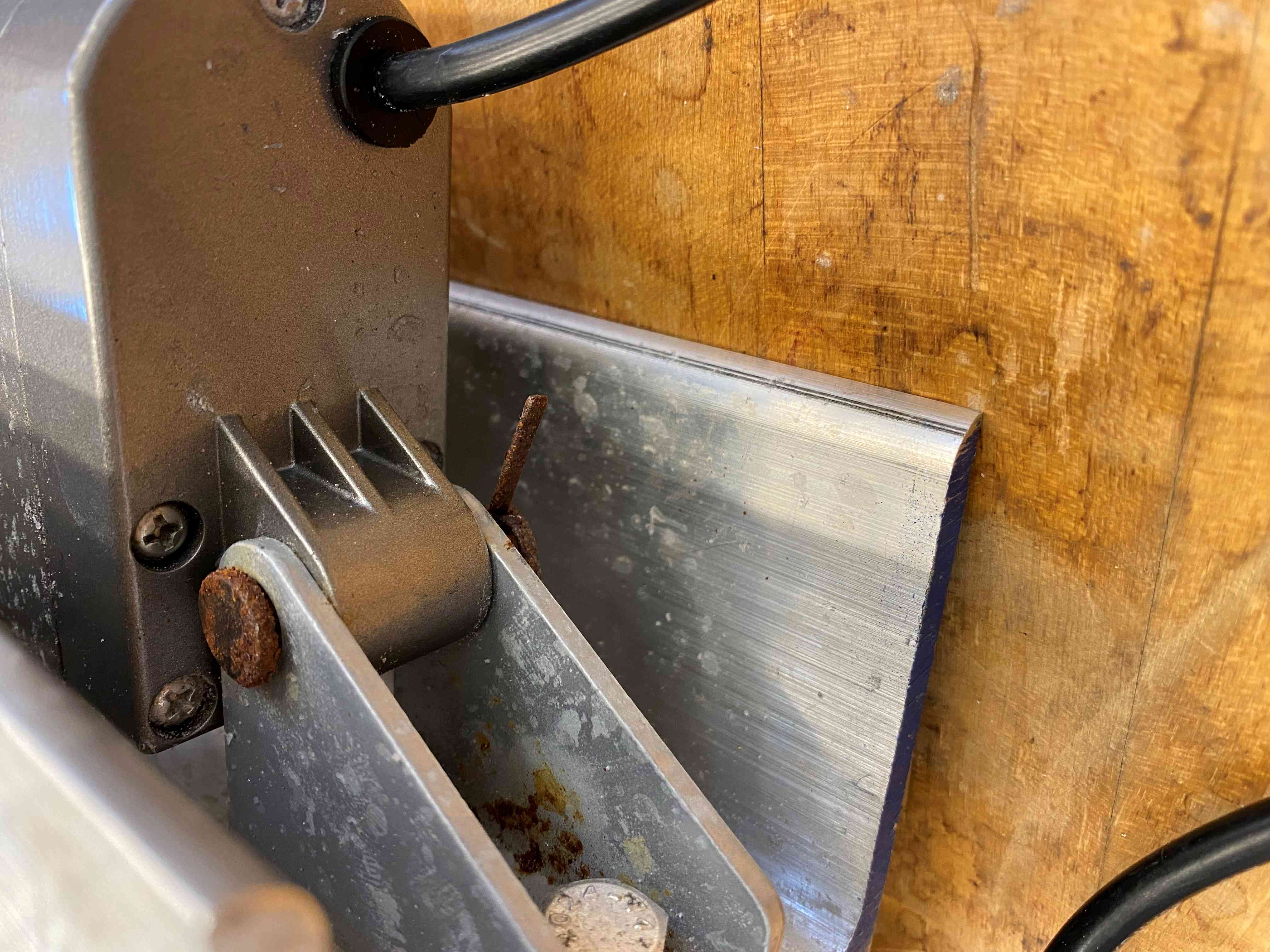}
\caption{}
\label{fig:corrosion3}
\end{subfigure}
\caption{(Color online) Corrosion after in-water testing of (RM)\textasciicircum2}
\label{fig:corrosion}
\end{figure*}

Generally, environmental conditions dictate the
external housing for maritime applications. Polymers and 
plastics perform the best in harsh salt-water conditions but lack the durability and robustness to handle the effects of sea state and varying towline tension considerations. These materials can be structurally weak when subjected to
external forces such as those involved in towing.
(RM)\textasciicircum2 was developed using a steel housing
which after only a few months of testing and exposure
to salt-water conditions, showed significant corrosion, as shown in Fig.~\ref{fig:corrosion}. The corrosion occurred
along the steel cotter pins that held the linear actuator in place and some oxidization occurred on parts
of the aluminum exterior.

Though we did not consider corrosion resistance as an immediate concern for rapid prototyping and testing, it is straightforward to implement corrosion mitigation through a more intelligent selection of housing materials and components. Stainless steel is known for its anti-corrosive properties, but it can be costly. Alternatively, there are corrosion inhibitor aerosols that can be purchased and applied to the housing directly to inhibit corrosion.
\section{Internet of Things Architecture Considerations}
\label{sec:iot_architecture_considerations}
With the proliferation of the Internet of Things (IoT) and advanced sensing technology, collecting and aggregating 
data across heterogeneous sensor networks has necessitated the development of architectures
and industry standards to ensure that sensors and computers
can communicate across different networking communication protocols,
hardware systems, and software systems. 

Three architectures have garnered support from the robotics community: Lightweight Communications and Marshalling (LCM), Data Distribution Service (DDS), and the Robot Operating System (ROS). 
Both LCM and DDS offer a publish-subscribe pattern that allows nodes to communicate directly, while still maintaining loose-coupling. LCM is an open-source set of tools that targets real-time systems that require high bandwidth (``fast") network speeds for delivering data. DDS is maintained by the Object Management Group (OMG). DDS also maintains loose coupling between nodes and has a vast user group that includes academic, 
None, however, has gained the widespread adaptation in robotics as ROS. ROS was developed at Stanford University, with a bulk of early development conducted by Willow Garage to create a generic framework for facilitating robot communication. It utilizes the publish-subscribe (see ``Design Patterns" by Gamma et al. for a precise description of this pattern \cite{gamma1994}) that relies on a ROS master node (which is a server) that communicates parameter updates to subscriber nodes.

\begin{figure}[h]
\centering
\includegraphics[width=1.0\columnwidth]{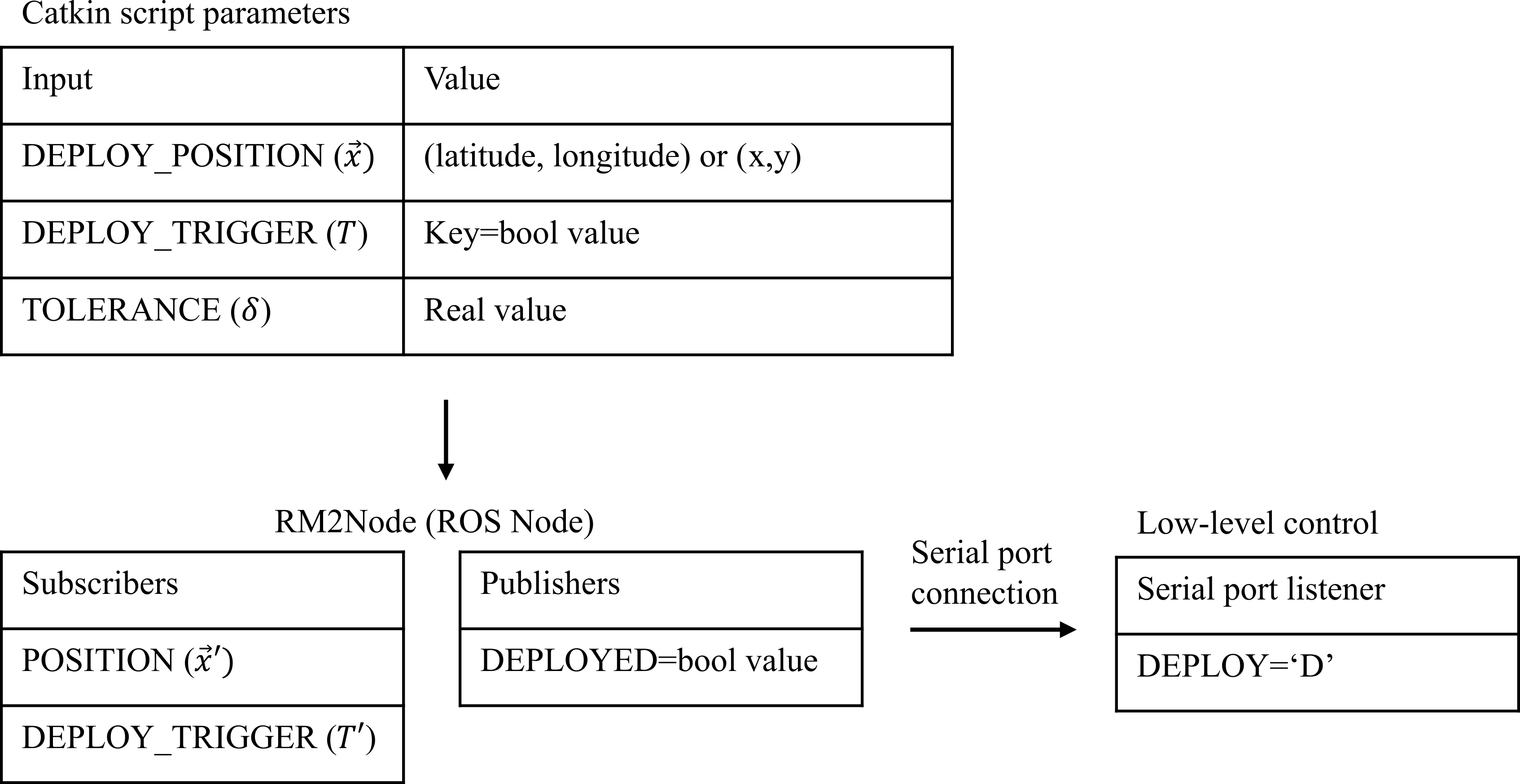}
\caption{Software architecture}
\label{fig:architecture}
\end{figure}

Since ROS has garnered significant attention in the robotics community, it was selected as the IoT protocol of choice for its widespread adaptation and utilization, particularly in maritime robotics applications. By developing the control software in ROS, (RM)\textasciicircum2 integrates with existing maritime autonomy software written in ROS. Fig.~\ref{fig:architecture} shows a high-level software architecture of (RM)\textasciicircum2. The $\textproc{RM2Node}$ utilizes a desired deployment location or deployment trigger to call $\textproc{WriteToSerialPort}$ which communicates with the low-level Arduino controller by sending a character that actuates the release mechanism. The algorithm is shown in Alg.~\ref{alg:rm2}.
	\begin{algorithm}[h]
\begin{algorithmic}[1]
\Procedure{RM2Node}{$\{\vec{x}~\textit{or}~\text{T}\}$,$\delta$}
\State $\vec{x}' = \textproc{ReceivePositionUpdate}()$
\State $\text{T}' = \textproc{ReceiveBoolTriggerUpdate}()$
\If{$\lvert\vec{x}'-\vec{x}\rvert < \delta~\text{or}~\text{T}'.\text{value}=\textit{true}$}
\State $\textproc{WriteToSerialPort}(\text{`D'})$
\EndIf
\State \Return $\textit{true}$
\EndProcedure
\end{algorithmic}
\caption{\label{alg:rm2} (RM)\textasciicircum2 algorithm}
\end{algorithm}

\section{A Theoretical Argument for Assured Release}
\label{sec:assured_release}
\begin{figure}[t]
\centering
\includegraphics[width=0.85\columnwidth]{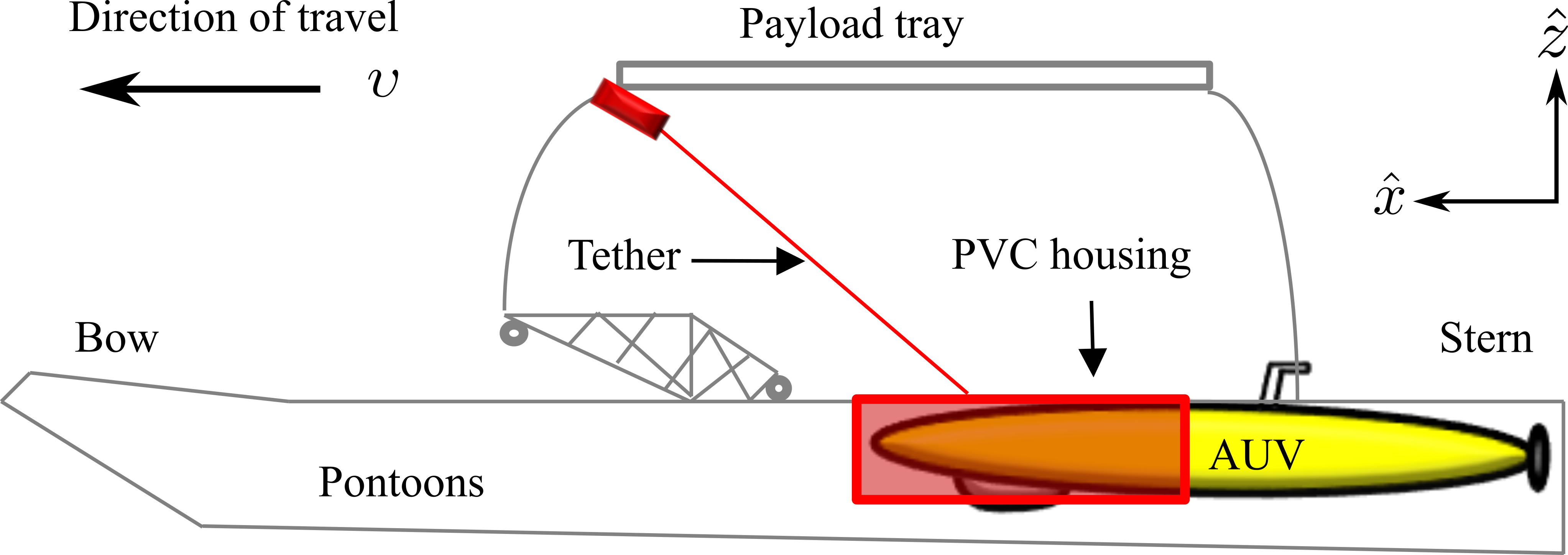}
\caption{Simplified representation of the (RM)\textasciicircum2 system. Drawing is not to scale}
\label{fig:simplified_schematic}
\end{figure}

We make an argument based on simplified assumptions that (RM)\textasciicircum2 will release under certain towing vessel velocity considerations. In Fig.~\ref{fig:simplified_schematic} we have drawn a representation of the AUV under tow. The towing vessel moves in the $\hat{x}$ direction with speed $\upsilon$. We have also shown an accompanying force body diagram on the center of mass of the AUV that includes all external forces acting on the AUV while under tow. Assuming that there is no net force acting on the AUV or any incident surface waves, then it is clear that the tension in the line implies (RM)\textasciicircum2 will always release. However, under non-zero surface wave conditions the towing vehicle must move with a speed of at least $A\omega$, where $A$ is the wave amplitude and $\omega$ is the angular frequency of the incident wave. To show this, we assume that an incident wave or plane progressive wave perturbs the towing system, causing both a heave and surge translation of the AUV in both $\hat{z}$ and $\hat{x}$ directions, respectively. These perturbations would cause slack in the towline, which would affect the ability of (RM)\textasciicircum2 to actually release the towed AUV. Under free surface assumptions, one can show that the resulting fluid velocity field is given by

\begin{equation}
\phi(x,z,t) = \frac{gA}{\omega}\text{e}^{kz}\text{sin}(kx-\omega t).
\label{eqn:potential}
\end{equation}

For a detailed derivation of (\ref{eqn:potential}) from wave mechanics, the interested reader should consult the work of Newman \cite{newman}. The velocity of the potential field in the $\hat{x}$ direction is given by $u = \partial \phi/\partial x$. Taking the derivative yields

\begin{equation}
u = A\omega\text{e}^{kz}\text{cos}(kx-\omega t).
\end{equation}

\noindent Notice that the maximum value of cosine is $1$. For the purposes of computing the surge velocity the coefficient of the cosine function dominates the surge velocity value. Also, assuming that $z << k = 2\pi/\lambda$ then effectively the leading order for the surge velocity is 

\begin{equation}
u \approx A\omega,
\end{equation}

\noindent where $\omega = 2\pi/\text{T}$, and $\text{T}$ is the period of the wave. Utilizing values for both wave amplitudes and wave periods that are encountered in a typical testing environment, $A = 0.50~\text{m}$ and a wind or gravity wave with a period of $\text{T} = 30~\text{s}$ then the towing vessel transit speed $\upsilon$ must be at least $0.105~\text{m/s}$ or about $0.20~\text{kn}$. The WAM-V vessel we used can achieve a maximum of $11~ \text{kn}$. Clearly, even at very slow operating conditions and moderate sea state conditions, then there is sufficient tension in the towline to guarantee release of the AUV.

\section{Experimental Testing}
\label{sec:experimental_testing}
Experimental testing of the (RM)\textasciicircum2 prototype was conducted in two parts: bench testing, to test simplistic assumptions about functionality and system performance, and in-water testing to verify system performance when deploying the AUV under realistic mission assumptions. 

\subsection{Bench Testing}
\label{sec:bench_testing}
\begin{figure}[t]
\centering
\includegraphics[width=0.55\columnwidth]{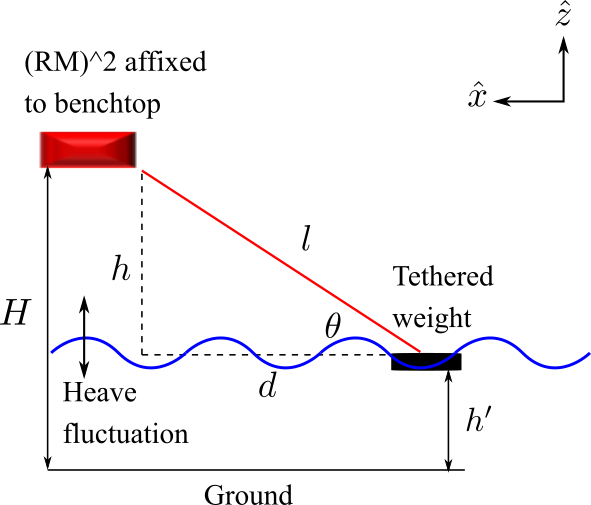}
\caption{(Color online) Bench test configuration for release performance under simulated heave conditions}
\label{fig:bench_test}
\end{figure}

Bench tests were conducted to ensure that (RM)\textasciicircum2 could accommodate different sea state conditions. Specifically, we created conditions in a laboratory setting to simulate both heave and surge translation. The experimental setup is shown in Fig.~\ref{fig:bench_test}. Heave conditions during an in-water test perturb the AUV in the $\hat{z}$-direction, perpendicular to the direction of travel. Effectively, this changes the angle between the vehicle measured along the equilibrium or steady sea condition and the tether. This angle is shown in Fig.~\ref{fig:bench_test}, labeled $\theta$. This angle is challenging to measure directly. Instead, we measured the height $h'$ or the distance between the floor and the testing weight, which was held at a controllable height. By varying the height $h'$, this effectively tests (RM)\textasciicircum2 under heave conditions. Care was taken to ensure that the line was always taut, as slack in the line would prevent (RM)\textasciicircum2 from releasing. This is not unsurprising, since any surge translation without a constant velocity in the $+\hat{x}$-direction results in slack and will not allow the device to release as expected. To keep the line taut, at each angle, the distance $d$ was increased. 

The bench test parameters are shown in Table~\ref{table:benchtop}. The distance between the WAM-V's payload tray and the steady state water line (the black solid line labeled ``Ground'') is equivalent to the bench height used in the laboratory or $H$ in the figure. Although in theory the AUV could fluctuate an arbitrary amount in the $\hat{z}$-direction, practically, this height is limited by the tether length that affixes the PVC housing to the pontoons. The housing tether lengths are $0.660~\text{m}$. This implies that AUV can only fluctuate with wave heights of approximately $1.32~\text{m}$. The keen observer will note that even this is practically limited by the tether length chosen as well, since at the lowest possible trough of $-0.660~\text{m}$ as measured from the mean water level, then $H+0.660=1.422~\text{m}$. This distance exceeds the length of the tether used as shown in the table. Hence, the AUV would effectively be suspended from the payload tray. To avoid this, we focused on sampling heave conditions that did not exceed the length of the tether or create unrealistic conditions like the AUV colliding with the payload tray of the WAM-V. In this case, wave amplitudes $h'$ between $-0.50~\text{m}$ and $0.50~\text{m}$ were scaled to match with a shortened tether to approximate the maximum and minimum angles produced in the real-time system. In the real-time system, the maximum angle that is possible occurs at the wave trough or $\theta_{\text{trough}}$, given by

\begin{equation}
\theta_{\text{trough}} = \sin^{-1}\bigg(\frac{H+h'_{\text{min}}}{l}\bigg).
\end{equation}

Conversely, the minimum angle occurs at wave crest, given by 

\begin{equation}
\theta_{\text{crest}} = \sin^{-1}\bigg(\frac{H-h'_{\text{max}}}{l}\bigg).
\end{equation}

These equations do not apply in the experimental case, since the origin does not occur at the ``Ground'' level shown in the figure. Instead, tether length was chosen so that the maximum and minimum angles could be approximated within the bench top height $H$. In the experimental case, 

\begin{equation}
\theta_{\text{trough,crest}} = \sin^{-1}\bigg(\frac{H-h'_{\text{min,max}}}{l}\bigg).
\end{equation}

In both instances of maximum and minimum angles, (RM)\textasciicircum2 successfully released the tether after five total iterations at each height. Since the device functioned as expected at both the maximum angle (the wave trough) and minimum angle (wave crest), then passing through equilibrium the device should release as well. 

\begin{table}[!t]
\renewcommand{\arraystretch}{1.1}
\caption{Real-time system dimensions and scaled experimental dimensions}
\label{table:benchtop}
\centering
\begin{tabular}{ccc}
\hline
Parameter&Real-time&Experimental\\
\hline\hline
tether length $l$&1.27~m&0.537~m\\
mean surface level $H$&0.762~m&0.762~m\\
wave trough depth $h'_{\text{min}}$&-0.50~m&0.152~m\\
wave crest height $h'_{\text{max}}$&0.50~m&0.532~m\\
$\theta_{\text{trough}}$&83.57\textdegree&83.57\textdegree\\
$\theta_{\text{crest}}$&16.66\textdegree&16.66\textdegree\\
\hline
\end{tabular}
\end{table}

\subsection{In-water Testing}
\begin{figure}[t]
\centering
\includegraphics[width=0.75\columnwidth]{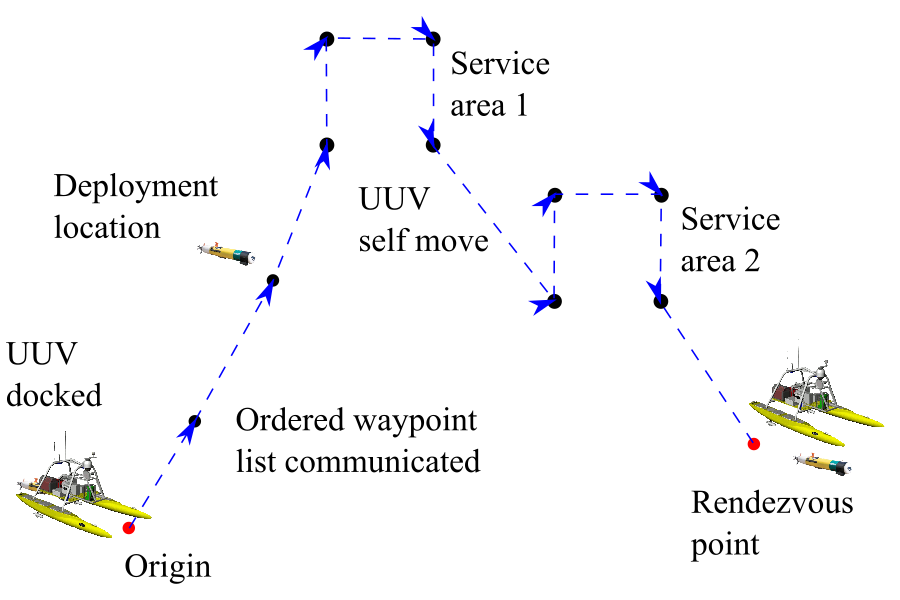}
\caption{(Color online) Mission schematic of the service agent transport problem under test}
\label{fig:mission}
\end{figure}

\begin{figure*}[t]
\centering
\begin{subfigure}[t]{0.49\textwidth}
\centering
\includegraphics[width=1.0\textwidth]{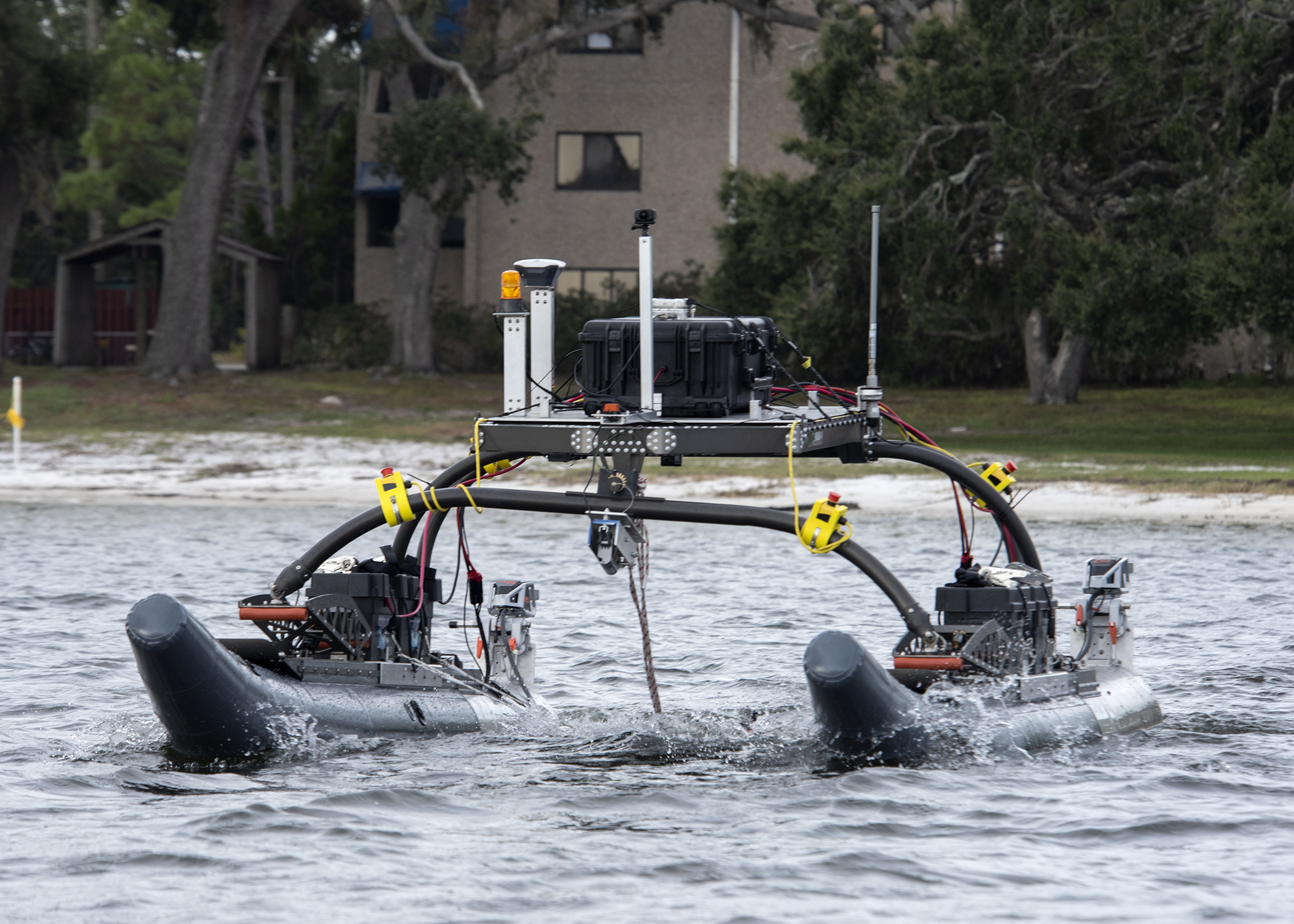}
\caption{Front view in-water test}
\label{fig:front_view_in_water_test}
\end{subfigure}
\begin{subfigure}[t]{0.49\textwidth}
\centering
\includegraphics[width=1.0\textwidth]{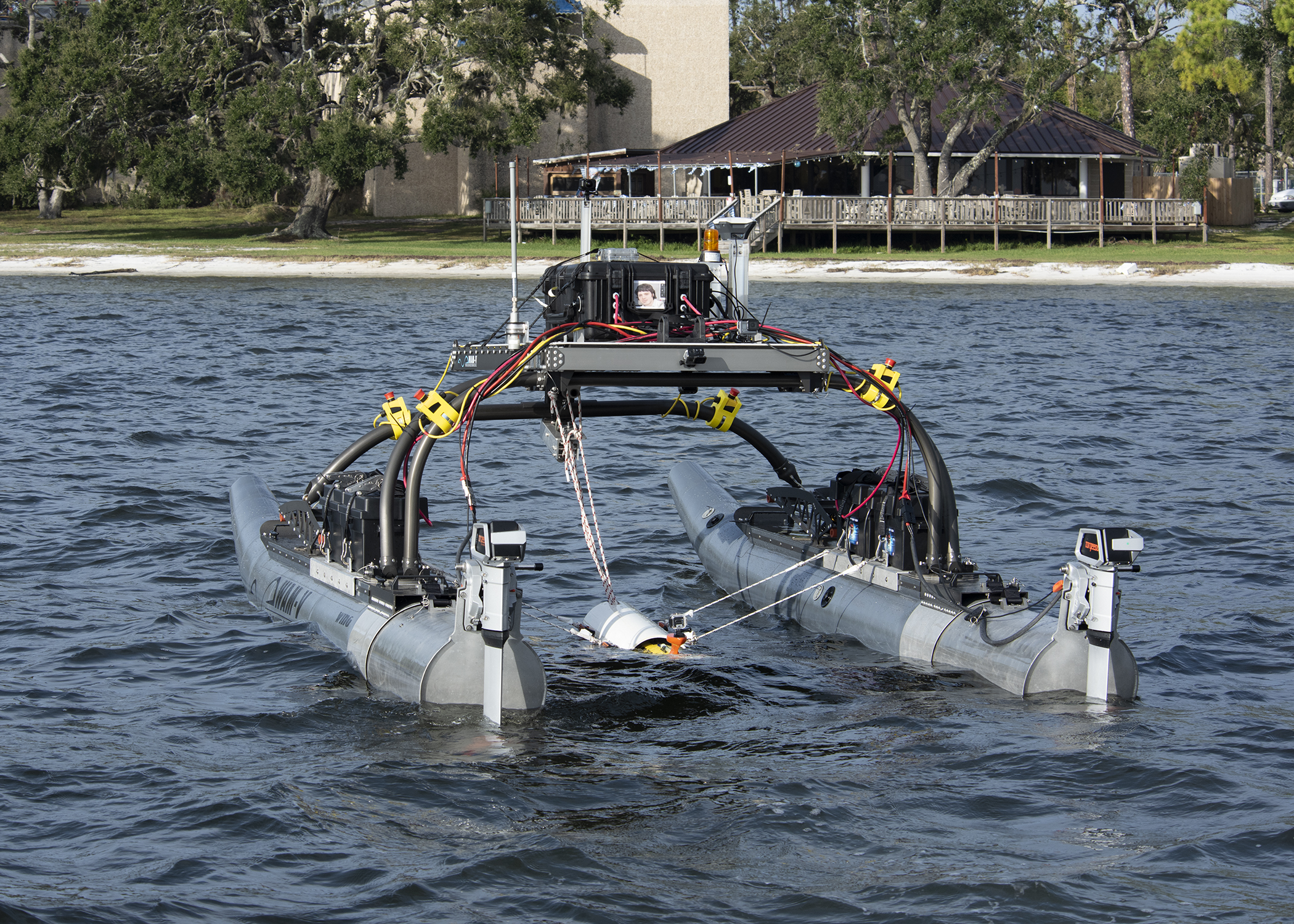}
\caption{Rear view in-water test}
\label{fig:rear_view_in_water_test}
\end{subfigure}
\caption{(Color online) In-water test images showing the AUV in the housing under tow by the WAM-V}
\label{fig:in_water_test}
\end{figure*}

In-water tests were conducted to determine how (RM)\textasciicircum2 performs when releasing a REMUS 100 from the WAM-V. Fig.~\ref{fig:in_water_test} shows the REMUS tethered to the WAM-V. To validate laboratory tests, the service agent transport problem (SATP) developed by Bays and Wettergren, was selected as a test case for autonomously triggering the deployment of the REMUS from the WAM-V \cite{Bays_SATPTaskAllocation,Bays_SATP,Bays_decoupled}. The SATP is a generalized scheduler for coordinating move, dock, deploy, and survey tasks across a team of ASVs and AUVs, where ASVs transport AUVs between survey areas. The algorithm utilizes a mixed-integer linear program to allocate tasks and near-optimally order waypoint transits. The mission schematic is shown in Fig.~\ref{fig:mission}. The WAM-V tows the AUV to a preset deployment location. En-route the WAM-V utilizes the SATP algorithm to optimize the order of the service areas and generate a `WAYPT\_UPDATE' publication in the form of an $x_1,y_1:\cdots: x_n,y_n$ string that the AUV parses and follows once deployed. This list is communicated over Ethernet to the AUV, utilizing a directional wireless antenna.  At the deployment location, the algorithm shown in Alg.~\ref{alg:rm2} executes and the AUV is deployed. At that point, the AUV follows the optimized waypoint list and rendezvouses with the WAM-V at a preset location. 

Ultimately, in-water tests proved fruitful for diagnosing a number of system issues. The list of optimized waypoints was generated and received by the REMUS vehicle; however, repeated testing showed that while the $\textproc{RM2Node}$ received both position updates and triggers to actuate, the device did not actuate. After many repeated attempts it was determined that a faulty actuator controller was to blame. After introducing the new actuator controller, repeated bench tests showed that (RM)\textasciicircum2 functioned as expected under simulated position and trigger updates. Ironically, funding levels did not permit extensive in-water testing after the first actuator controller malfunction.

\section{Conclusion}
\label{sec:conclusion}
We have outlined the development of a remotely-enabled
modular release mechanism called (RM)\textasciicircum2 that facilitates
the autonomous release of an autonomous underwater vehicle
from catamaran-like vehicles. While we have conducted bench tests
and sea tests utilizing the WAM-V catamaran, (RM)\textasciicircum2 is adaptable 
to a variety of different vehicle types and is not restricted to any specific
vehicle implementation. We have shown that through careful
design considerations (RM)\textasciicircum2 is suitable for a variety of different
applications via the standardized ROS interface built to control the 
low-level actuation hardware. Since it is easily assembled and low-cost,
(RM)\textasciicircum2 is suitable for early-stage and proof of concept autonomy 
projects that require cooperative autonomy between ASV and AUVs. 

Future development will look at a feedback mechanism to inform the autonomy software of release which differentiates between success and failure. This will help diagnose system errors and provide a better troubleshooting mechanism for developers.
\section*{Declarations}
D. T. Kutzke, G. E. Miranda L\'{o}pez, and R. J. Herman declare that
they have submitted a patent application to the United States Patent and Trademark
Office to cover the release mechanism described in the paper. H. Philippeaux 
has no financial conflicts of interests or competing interests to report.

\bibliographystyle{abbrv}
\bibliography{Bibliography}

\begin{thebibliography}{10}

\bibitem{ackerman2018medical}
E.~Ackerman and E.~Strickland.
\newblock Medical delivery drones take flight in east africa.
\newblock {\em IEEE Spectr.}, 55(1):34--35, 2018.

\bibitem{allen2006autonomous}
B.~Allen, T.~Austin, N.~Forrester, R.~Goldsborough, A.~Kukulya, G.~Packard, M.~Purcell, and R.~Stokey.
\newblock Autonomous docking demonstrations with enhanced {REMUS} technology.
\newblock In {\em MTS/IEEE OCEANS 2006}, pages 1--6. IEEE, 2006.

\bibitem{anderlini2019_DockingControlAUVReinforcement}
E.~Anderlini, G.~G. Parker, and G.~Thomas.
\newblock Docking control of an autonomous underwater vehicle using reinforcement learning.
\newblock {\em Appl. Sci.}, 9(17), 2019.

\bibitem{ansay2008pre}
M.~T. Ansay and A.~Di~Biasio.
\newblock Pre-positioning deployment system, Sept.~2 2008.
\newblock {US} Patent 7,418,914.

\bibitem{banerjee1995deployment}
A.~K. Banerjee and V.~N. Do.
\newblock Deployment control of a cable connecting a ship to an underwater vehicle.
\newblock {\em Appl. Math. Comput.}, 70(2-3):97--116, 1995.

\bibitem{Bays_SATP}
M.~J. Bays and T.~A. Wettergren.
\newblock A solution to the service agent transport problem.
\newblock In {\em 2015 {IEEE/RSJ IROS}}, pages 6443--6450. IEEE, September 2015.

\bibitem{Bays_SATPTaskAllocation}
M.~J. Bays and T.~A. Wettergren.
\newblock Service agent-transport agent task planning incorporating robust scheduling techniques.
\newblock {\em Robot. Auton. Syst.}, 89:15--26, March 2017.

\bibitem{Bays_decoupled}
M.~J. Bays and T.~A. Wettergren.
\newblock Partially-decoupled service agent-transport agent task allocation and scheduling.
\newblock {\em J. Intell. Robot. Syst.}, 94:423--437, April 2018.

\bibitem{birk2011}
A.~Birk, G.~Antonelli, A.~Caiti, G.~Casalino, G.~Indiveri, A.~Pascoal, and A.~Caffaz.
\newblock The co<sup>3</sup>auvs (cooperative cognitive control for autonomous underwater vehicles) project: Overview and current progresses.
\newblock In {\em OCEANS 2011 IEEE - Spain}, pages 1--10, 2011.

\bibitem{chen2019cost}
Y.~Chen, X.~Mao, S.~Yang, and Q.~Wang.
\newblock Cost-efficient inter-robot delivery for resource-constrained and interdependent multi-robot schedules.
\newblock {\em Int. J. Adv. Robot. Syst.}, 16(1):1--26, 2019.

\bibitem{ead2007apparatus}
R.~M. Ead and R.~L. Pendleton.
\newblock Apparatus for deploying and recovering a towed acoustic line array from an unmanned undersea vehicle, Aug.~7 2007.
\newblock {US} Patent 7,252,046.

\bibitem{fiaz2018intelligent}
U.~A. Fiaz, M.~Abdelkader, and J.~S. Shamma.
\newblock An intelligent gripper design for autonomous aerial transport with passive magnetic grasping and dual-impulsive release.
\newblock In {\em 2018 IEEE/ASME Int. Conf. AIM}, pages 1027--1032, Auckland, New Zealand, 2018.

\bibitem{gamma1994}
E.~Gamma, R.~Helm, R.~Johnson, J.~Vlissides, and G.~Booch.
\newblock {\em Design Patterns: {E}lements of Reusable Object-Oriented Software}.
\newblock Addison-Wesley Professional, 1 edition, 1994.

\bibitem{Gatteschi2015_DroneDeliverySystems}
V.~{Gatteschi}, F.~{Lamberti}, G.~{Paravati}, A.~{Sanna}, C.~{Demartini}, A.~{Lisanti}, and G.~{Venezia}.
\newblock New frontiers of delivery services using drones: {A} prototype system exploiting a quadcopter for autonomous drug shipments.
\newblock In {\em 2015 IEEE 39th Annu. Comput. Softw. Appl. Conf.}, volume~2, pages 920--927, Taichung, Taiwan, 2015.

\bibitem{gerkey_taxonomy}
B.~P. Gerkey and M.~J. Matari{\'c}.
\newblock A formal analysis and taxonomy of task allocation in multi-robot systems.
\newblock {\em Int. J. Robot. Res.}, 23(9):939--954, September 2004.

\bibitem{gonzalez2020_AUVCollaboration}
J.~González-García, A.~Gómez-Espinosa, E.~Cuan-Urquizo, L.~G. García-Valdovinos, T.~Salgado-Jiménez, and J.~A.~E. Cabello.
\newblock Autonomous underwater vehicles: {L}ocalization, navigation, and communication for collaborative missions.
\newblock {\em Appl. Sci.}, 10(4), 2020.

\bibitem{hobson2007development}
B.~W. Hobson, R.~S. McEwen, J.~Erickson, T.~Hoover, L.~McBride, F.~Shane, and J.~G. Bellingham.
\newblock The development and ocean testing of an {AUV} docking station for a 21" {AUV}.
\newblock In {\em IEEE/MTS OCEANS 2007}, pages 1--6. IEEE, 2007.

\bibitem{jorge2019survey}
V.~A. Jorge, R.~Granada, R.~G. Maidana, D.~A. Jurak, G.~Heck, A.~P. Negreiros, D.~H. dos Santos, L.~M. Gon{\c{c}}alves, and A.~M. Amory.
\newblock A survey on unmanned surface vehicles for disaster robotics: {M}ain challenges and directions.
\newblock {\em Sensors}, 19(3):702, 2019.

\bibitem{Kalwa2009}
J.~Kalwa.
\newblock The grex-project: Coordination and control of cooperating heterogeneous unmanned systems in uncertain environments.
\newblock In {\em {OCEANS} 2009-{EUROPE}}, pages 1--9, 2009.

\bibitem{Kalwa2015}
J.~Kalwa, A.~Pascoal, P.~Ridao, A.~Birk, T.~Glotzbach, L.~Brignone, M.~Bibuli, J.~Alves, and M.~Silva.
\newblock {EU} project {MORPH}: Current status after 3 years of cooperaton under and above water.
\newblock {\em IFAC NGCUV}, 48(2), 2015.

\bibitem{korsah2012xbots}
G.~A. Korsah, B.~Kannan, B.~Browning, A.~Stentz, and M.~B. Dias.
\newblock {xBots}: {A}n approach to generating and executing optimal multi-robot plans with cross-schedule dependencies.
\newblock In {\em 2012 IEEE ICRA}, pages 115--122, Saint Paul, MN, USA, 2012. IEEE.

\bibitem{kouriampalis2021_OperationalEffects}
N.~Kouriampalis, R.~Pawling, and D.~Andrews.
\newblock Modelling the operational effects of deploying and retrieving a fleet of uninhabited vehicles on the design of dedicated naval surface ships.
\newblock {\em Ocean Eng.}, 219:108274, 2021.

\bibitem{kutzke2021generosity}
D.~T. Kutzke, R.~D. Tatum, and M.~J. Bays.
\newblock Generosity-based schedule deconfliction in communication-limited environments.
\newblock {\em J. Intell. Robot. Syst.}, 101(1):1--20, 2021.

\bibitem{Landstad2021_DynamicPositioning}
O.~Landstad, H.~S. Halvorsen, H.~Øveraas, V.~Smines, and T.~A. Johansen.
\newblock Dynamic positioning of {ROV} in the wave zone during launch and recovery from a small surface vessel.
\newblock {\em Ocean Eng.}, 235:109382, 2021.

\bibitem{limosani2018robotic}
R.~Limosani, R.~Esposito, A.~Manzi, G.~Teti, F.~Cavallo, and P.~Dario.
\newblock Robotic delivery service in combined outdoor--indoor environments: technical analysis and user evaluation.
\newblock {\em Robot. Auton. Syst.}, 103:56--67, 2018.

\bibitem{yukun2018_UnderwaterVehicleForMarineLife}
Y.~Lin, J.~Hsiung, R.~Piersall, C.~White, C.~G. Lowe, and C.~M. Clark.
\newblock A multi-autonomous underwater vehicle system for autonomous tracking of marine life.
\newblock {\em J. Field Robot.}, 34(4):757--774, 2017.

\bibitem{link2010government}
A.~N. Link and J.~T. Scott.
\newblock Government as entrepreneur: Evaluating the commercialization success of {SBIR} projects.
\newblock {\em Res. Policy}, 39(5):589--601, 2010.

\bibitem{wamv}
Marine Advanced Robotics, Inc, Richmond, CA.
\newblock {\em {WAM-V} 16 {ASV}}, 2021.

\bibitem{meng2019_UnderwaterDocking}
L.~Meng, Y.~Lin, H.~Gu, and T.-C. Su.
\newblock Study on dynamic docking process and collision problems of captured-rod docking method.
\newblock {\em Ocean Eng.}, 193:106624, 2019.

\bibitem{newman}
J.~Newman.
\newblock {\em Marine Hydrodynamics}.
\newblock The MIT Press, 1977.

\bibitem{moosgeodesy}
P.~Newman.
\newblock {\em {libMOOSGeodesy}}.
\newblock moos-ivp.org, 2016.

\bibitem{patrick2016mechanisms}
W.~G. Patrick, J.~R. Burgess, and A.~Conrad.
\newblock Mechanisms for lowering a payload to the ground from a uav, May~24 2016.
\newblock {US} Patent 9,346,547.

\bibitem{pugi2018redundant}
L.~Pugi, B.~Allotta, and M.~Pagliai.
\newblock Redundant and reconfigurable propulsion systems to improve motion capability of underwater vehicles.
\newblock {\em Ocean Eng.}, 148:376--385, 2018.

\bibitem{real2016}
D.~A. Real-Arce, T.~Morales, C.~Barrera, J.~Hern{\'a}ndez, and O.~Llin{\'a}s.
\newblock Smart and networking underwater robots in cooperation meshes - the swarms ecsel-h2020 project.
\newblock In {\em 7th Int. Workshop Marine Technol.}, Barcelona, Spain, 26--28 Oct. 2016.

\bibitem{renilson2014simplified}
M.~Renilson.
\newblock A simplified concept for recovering a {UUV} to a submarine.
\newblock {\em Underw. Technol.}, 32(3):193--197, 2014.

\bibitem{rizk2019cooperative}
Y.~Rizk, M.~Awad, and E.~W. Tunstel.
\newblock Cooperative heterogeneous multi-robot systems: {A} survey.
\newblock {\em ACM Comput. Surv.}, 52(2):1--31, 2019.

\bibitem{sarda2014}
E.~Sarda, M.~Dhanak, and K.~Ellenrieder.
\newblock A {USV}-based automated launch and recovery system for {AUVs}.
\newblock {\em IEEE J. Ocean. Eng.}, 42(1):37--55, 2016.

\bibitem{song2020care}
J.~Song and S.~Gupta.
\newblock {CARE}: Cooperative autonomy for resilience and efficiency of robot teams for complete coverage of unknown environments under robot failures.
\newblock {\em Auton. Robot.}, 44(3):647--671, 2020.

\bibitem{song2013adaptive}
J.~Song, S.~Gupta, J.~Hare, and S.~Zhou.
\newblock Adaptive cleaning of oil spills by autonomous vehicles under partial information.
\newblock In {\em 2013 MTS/IEEE OCEANS}, pages 1--5, San Diego, CA, USA, 2013. IEEE.

\bibitem{cris2021_TaskPriority}
C.~Thomas, E.~Simetti, and G.~Casalino.
\newblock A unifying task priority approach for autonomous underwater vehicles integrating homing and docking maneuvers.
\newblock {\em J. Mar. Sci. Eng.}, 9(2), 2021.

\bibitem{trslic2020neuro}
P.~Trsli{\'c}, E.~Omerdic, G.~Dooly, and D.~Toal.
\newblock Neuro-fuzzy dynamic position prediction for autonomous work-class {ROV} docking.
\newblock {\em Sens.}, 20(3):693, 2020.

\bibitem{utne2019}
I.~B. Utne, I.~Schjølberg, and E.~Roe.
\newblock High reliability management and control operator risks in autonomous marine systems and operations.
\newblock {\em Ocean Eng.}, 171:399--416, 2019.

\bibitem{wigley2018}
R.~Wigley, A.~A. Proctor, and B.~Simpson.
\newblock Novel {AUV} launch, recovery new approaches using combined {USV}-{AUV} method.
\newblock {\em Sea. Techn.}, 59(6):24--27, 2018.

\bibitem{zhang2021_Hydro}
W.~Zhang, G.~Jia, P.~Wu, S.~Yang, B.~Huang, and D.~Wu.
\newblock Study on hydrodynamic characteristics of {AUV} launch process from a launch tube.
\newblock {\em Ocean Eng.}, 232:109171, 2021.

\bibitem{zhao2021_LRD}
C.~Zhao, P.~Thies, J.~Lars, and J.~Cowles.
\newblock {ROV} launch and recovery from an unmanned autonomous surface vessel – hydrodynamic modelling and system integration.
\newblock {\em Ocean Eng.}, 232:109019, 2021.

\bibitem{zitouni2020towards}
F.~Zitouni, S.~Harous, and R.~Maamri.
\newblock Towards a distributed solution to the multi-robot task allocation problem with energetic and spatiotemporal constraints.
\newblock {\em Comput. Sci.}, 21(1), 2020.

\end{thebibliography}
\end{document}